\pdfoutput=1

\documentclass[11pt]{article}
\usepackage{authblk}
\usepackage{EMNLP2022}

\usepackage{times}
\usepackage{latexsym}

\usepackage[T1]{fontenc}

\usepackage[utf8]{inputenc}

\usepackage{microtype}

\usepackage{inconsolata}
\usepackage{graphicx}
\usepackage{amsmath}
\usepackage{booktabs}
\usepackage{algorithm}
\usepackage{algorithmic}
\usepackage{enumerate}
\usepackage{makecell}
\usepackage{hyperref} 
\usepackage{lipsum}
\usepackage{color}
\usepackage{color}
\usepackage{amssymb}
\usepackage{times}
\usepackage{latexsym}
\usepackage{pifont}
\usepackage{makecell}
\usepackage{float}
\usepackage{graphicx}
\usepackage{amssymb}
\usepackage{color}
\usepackage{amsmath}
\usepackage{multirow}
\newcommand{\xmark}{\ding{55}}%

\usepackage{blindtext}
\usepackage{microtype}
\usepackage[misc]{ifsym}
\title{Analogical Math Word Problems Solving with Enhanced Problem-Solution Association}

\author[1]{\textbf{Zhenwen Liang}}
\author[2]{\textbf{Jipeng Zhang}}
\author[1]{\textbf{Xiangliang Zhang}\textsuperscript{\tiny \Letter}}
\affil[1]{University of Notre Dame, \texttt{\;\{zliang6, xzhang33\}@nd.edu}}
\affil[2]{Hong Kong University of Science and Technology, \texttt {jzhanggr@conect.ust.hk}}

\begin{document}
\maketitle
\begin{abstract}
Math word problem (MWP) solving is an important task in question answering which requires human-like reasoning ability. Analogical reasoning  has long been used in mathematical education, as it enables students to apply common relational structures of mathematical situations to  solve new problems. In this paper, we propose to build a novel MWP solver by leveraging analogical MWPs, which advance the solver's generalization ability across different kinds of MWPs. The key idea, named  \textit{analogy identification},  is to associate the analogical MWP pairs in a latent space, i.e.,  encoding an MWP close to another analogical MWP, while moving away from the non-analogical ones. Moreover, a \textit{solution discriminator} is integrated into the MWP solver to enhance the association between the representations of MWPs and their true solutions. The evaluation results verify that our proposed analogical learning strategy promotes the performance of \textit{MWP-BERT} on Math23k over  the state-of-the-art model \textit{Generate2Rank}, with 5 times fewer parameters in the encoder. We also find that our model has a stronger generalization ability in solving  difficult MWPs due to the analogical learning from easy MWPs. 


\end{abstract}

\section{Introduction}

\begin{figure}
\centering 
\includegraphics[width=0.48\textwidth]{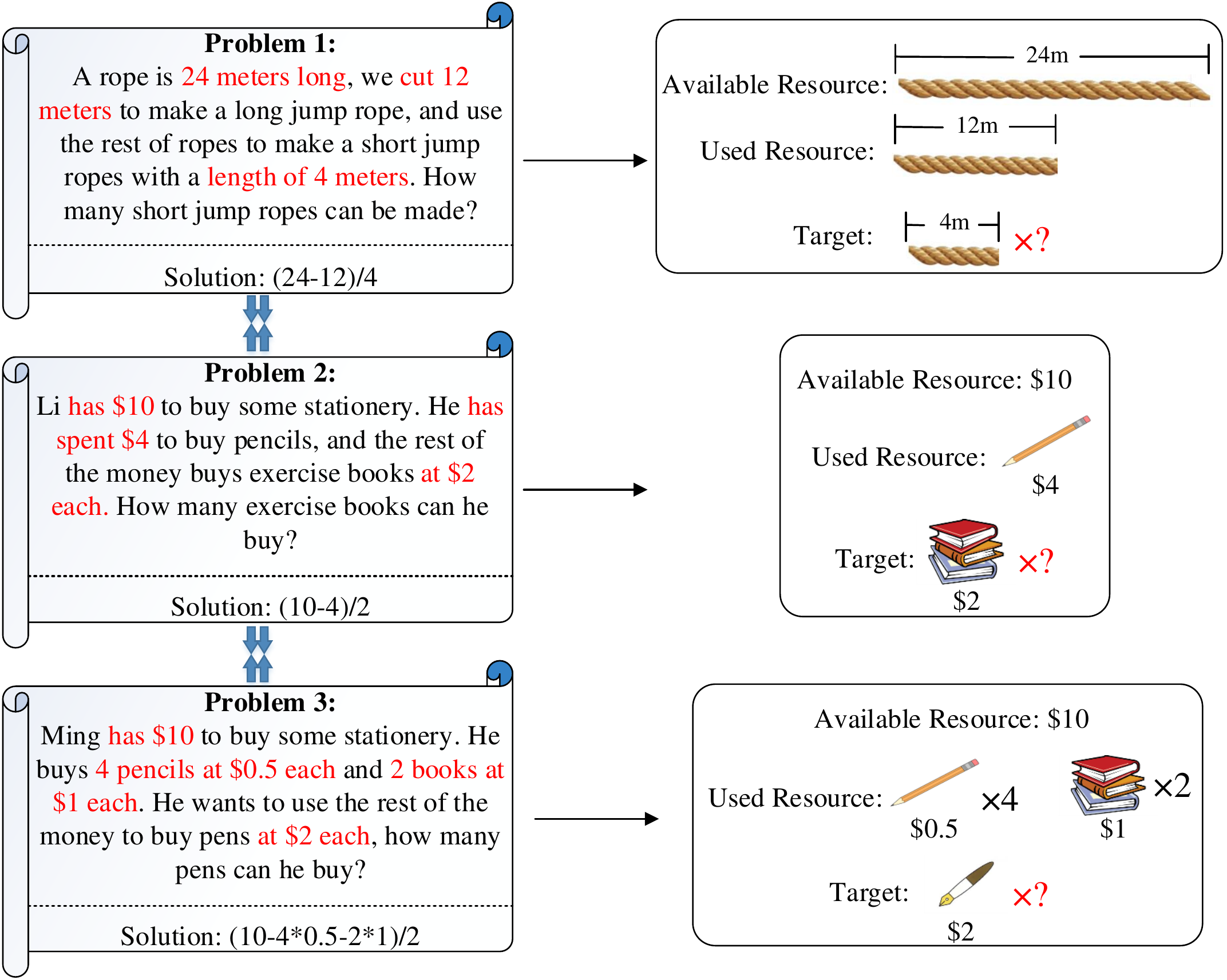}
\caption{This figure shows three examples of math word problems. We can find mathematical analogy across different topics (P1 and P2) and different difficulty levels (P2 and P3).}
\label{fig:intro} 
\end{figure}

Math word problem (MWP) solving has attracted considerable attention in recent years. Currently, MWP solver design focuses on generating an equation towards an unknown quantity, with an input problem description, as shown in Figure \ref{fig:intro}. 
Building such a successful solver is quite challenging, as it requires mathematical understanding and multi-step reasoning abilities to transform the implied logic behind the problem into a mathematical equation composed of operators and quantities. With the emergence of deep learning, MWP has been effectively studied by Seq2Seq models \cite{xie2019goal,zhang2020graph} and pre-trained language models (PLMs)  \cite{tan2021investigating,li2021seeking,huang2021recall,liang2022mwp}, which treat it as a translation task from the natural language description of a problem to its solution in a mathematical equation.

In educational science, analogical reasoning \cite{novick1991mathematical,bernardo2001analogical} has long been considered as a crucial skill in human problem-solving. In education and training of mathematical reasoning,   researchers \cite{stacey1982thinking} interpret it as ``a dynamics process which, by enabling us to increase the complexity of ideas we can handle, expands our understanding.'' For example, the MWP P1 and P2 in Figure \ref{fig:intro} are in different situations, but both aim to calculate the available quantity of a target object based on known resources, leading to the same solution structure. The MWP P2 and P3  both focus on the knowledge of  `number of units = money / unit price', while the MWP P3 has more reasoning steps.
Gaining the skill of generating the solution of P2 to address P1 and P3 is important for students, as well as for automated MWP solvers.
A widely accepted definition of analogical reasoning is mapping the process of mathematization between different domains \cite{vosniadou1988analogical}. 
We thus propose  a strategy of associating the mathematization process of analogical MWP pairs, e.g., P1 and P2, P2 and P3. The key idea is to encode the analogical problems   close in the latent representation space, while encoding the non-analogical problems distantly. 
The introduction of  analogical learning in MWP solving can then alleviate the limitations of existing methods such as simple-to-difficult generalization \cite{zhang2020graph,jie2022learning} and topic-to-topic generalization \cite{hong2021smart}.



While the analogical MWPs have close representations, they should be also more associated with their corresponding true solutions than wrong solutions. We thus encode the solution as well, and design a  module named solution discrimination to enhance the association of an MWP and its correct solution. The enhancement is powered by generating hard negative samples that are wrong solutions but similar to the correct solution. 
Negative sampling has been proved effective in training strong MWP solvers \cite{li2021seeking,ijcai2021}. For example, \citet{li2021seeking} used negative examples to calculate a contrastive loss and find patterns across MWPs, but the negative examples are retrieved from the given training set, which could be bounded by the dataset size and quality. \citet{ijcai2021} bridged the problem space and solution space, by building a teacher module to help the solver match problems with their ground truth solutions and distinguish them from randomly generated negative solutions. 
However, the negative samples in their approach are randomly manipulated variants of the ground truth equation, and are thus limited in their shapes and fail to consider the hardness of negative samples. Therefore, we design a gradient-based manipulation method to find hard negative samples (i.e., wrong solutions that are difficult  to be distinguished from a correct solution). In the meantime, to make the form of negative samples less bounded by the ground truth, we introduce more randomness during the generation of negative samples.

Overall, our proposed MWP solver is novel on 
considering the analogy among MWPs. It contains an \textit{analogy identification} module which aims to transfer mathematization knowledge and  skills to generalize the solution   to analogical MWPs. Moreover, a \textit{solution discrimination} module is incorporated to enable the solver to bridge problems and their ground truth solutions. Our proposed solver is a flexible framework and can have any MWP encoder-decoder plugged in. 
The experimental results show that training  
representative baselines (e.g., GTS \cite{xie2019goal} and MWP-BERT \cite{liang2022mwp}) by our analogical strategy can further improve their accuracy.  
\textcolor{black}{Comparing to the current best solver on Math23k, Generate\&Rank  \cite{shen2021generate}, we make MWP-BERT achieve higher accuracy but with 5 times fewer encoder parameters. }  
Ablation studies, qualitative analyses and case studies also demonstrate the effectiveness of our designed solver framework, and verify its generalization
 ability in solving difficult MWPs due to the analogical learning from easy MWPs.

\section{Related Works}
\subsection{Math Word Problem Solving}
The topic of automated math word problem solving was raised in 1960s \cite{bobrow1964natural}. At the beginning, researchers  established rules and tried to match every problem with a certain solving rule \cite{bakman2007robust}. Then, statistical and machine learning methods \cite{shi2015automatically} were widely applied and sometimes integrated with semantic parsing approaches \cite{koncel2015parsing}. Inspired by the great success of deep learning and Seq2Seq models, a Seq2Seq solver with an encoder-decoder structure was proposed and outperformed transitional methods, after which many other Seq2Seq solvers \cite{wang2018translating,wang2019template,liu2019tree} were proposed. Later, GTS \cite{xie2019goal} replaced the sequential decoder with a novel tree-based decoder and achieved great performance. Most following papers \cite{zhang2020graph,wu2020knowledge,lin2021hms,wu2021edge,ijcai2021} after GTS focused on the improving of encoder part without touching  the tree-based decoder. Recently, due to the development of pre-trained language models (PLMs) \cite{devlin2019bert,cui2020revisiting}, the accuracy of MWP-BERT \cite{liang2022mwp} firstly surpasses human performance \cite{wang2019template}. Besides accuracy improvement, there are many works exploring some other aspects of MWP solvers, e.g., teacher-student distillation  \cite{ijcai2020-555}, auxiliary training tasks  \cite{qin2021neural}, situation model  \cite{hong2021smart}. 
For more comprehensive reviews of MWP solver, please refer to these surveys \cite{zhang2019gap,faldu2021towards}.

\subsection{Analogical Learning in NLP}
The k-nearest neighbor retrieval has been a popular approach deployed in machine translation \cite{he2021efficient,jiang2021learning} since the born of kNN-MT \cite{khandelwal2021nearest}. It is simple yet effective in terms of improving the translation performance and explainability. This idea was quickly adopted by other NLP research such as building language models \cite{liu2022relational} and text-generation \cite{li2022survey}. There is one recent study in MWP solving that shows the benefits of retrieval-based analogical learning \cite{huang2021recall}. 
The basic idea is to set up a memory module for storing the problems that have been learned. For a new given problem, the solver retrieves \textcolor{black}{similar problems from the memory based on measuring the cosine similarities of the encoded problems and lets the solution of the retrieved problem participate  decoding.}
However, this approach has several limitations. Firstly, the retrieved problems with high cosine similarity  may not be analogical to the given problem. They have similar representation vectors perhaps  because they have common topic words in the problem description, but their reasoning steps are completely different.  These literally similar but non-analogical
problems can mislead the decoder   give a wrong answer. 
Secondly, with the increasing size of the training dataset/memory module, the solver will suffer from long search time. 

\textcolor{black}{
We design our analogical reasoning solver with an \emph{analogy identification} module, which takes advantage of only the  problems whose mathematical analogy is confirmed by their solution. Therefore, we have no  top-$k$ selection issues. Moreover, unlike \cite{huang2021recall} that introduces the analogical problems in decoding, we use them for improving  problem understanding (encoding). Analogical problem pairs are mapped close in a dense region, and  non-analogical pairs are separated in different regions (as demonstrated later in Figure 3). This adjusted representation space facilitates the decoding process and leads to more correct results.  
}

\section{Approach}

\begin{figure*}
\centering 
\includegraphics[width=1\textwidth]{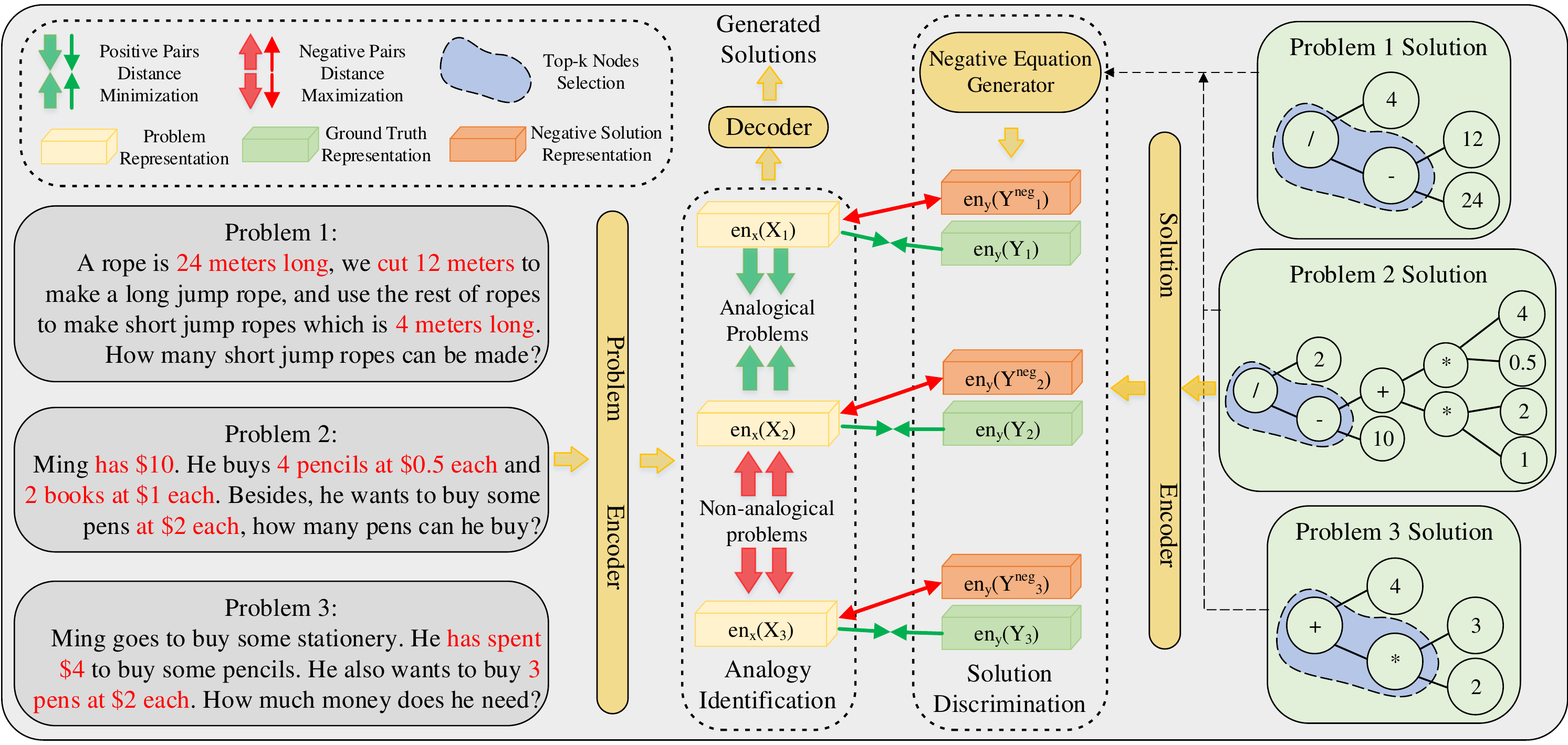}
\caption{The figure shows the workflow of our proposed framework with 3 example problems and solutions. P1 and P2 have the same top-1 and top-2 nodes, thus they are positive pairs in the analogy identification module, while P2 and P3 is a negative pair. In the solution discrimination part, each problem representation is paired with their ground truth solution to minimize their distance. Then, negative solutions are generated by the generator and the distance between them and the problem representation will be maximized.}
\label{fig:main} 
\end{figure*}

\subsection{Problem Definition}
We aims to train a math word problem solver that receives the problem description $X$ as input and then generates an equation-shaped solution $Y = \{y_1,y_2,...,y_n\}$ with length $n$. All equation solutions are transformed into a binary tree representation as proposed in  \cite{xie2019goal} and sequantialized as their pre-order traversal, thus there exists no parenthesis in $Y$. The vocabulary of $Y$ contains two parts, {operators and numbers}.
In the solution tree, operator nodes are always parents of numbers, and the number nodes have to be leaf nodes, as shown in Figure \ref{fig:main}. Specifically, the vocabulary of operators $V_{op}$ contains 5 operators $\{+,-,*,/,\string^\}$ and the vocabulary of numbers $V_{num}$ may vary across different problems. We apply number mapping  \cite{wang2017deep} to replace the numbers in $Y$ which can be found in $X$ with placeholders, thus the length of $V_{num}$ depends on how many numbers  appear in $X$. Besides, there are some useful constants like $\pi$ which are not shown in $X$ but used in $Y$ to get a solution, we also include them in $V_{num}$.

\subsection{Backbone Solvers}
\textcolor{black}{Our target is to design a model that can solve MWP by analogical reasoning. Therefore, the model is in fact  a flexible framework that can be implemented with any existing MWP solvers.}
We select GTS \cite{xie2019goal} and MWP-BERT \cite{liang2022mwp} as backbone solvers and plug them  into our novel framework. GTS uses GRU \cite{cho2014learning} as the encoder and MWP-BERT has a BERT encoder that is continually pre-trained on MWP corpus. Both of them apply tree-based decoder and two models can be found at open-sourced\footnote{\url{https://github.com/ShichaoSun/math_seq2tree} and \url{https://github.com/LZhenwen/MWP-BERT}}.

\subsection{Analogy Identification}
School students are often taught with example problems and tested by analogical problems in exams for evaluating their problem-solving skills. As stated in \cite{gentner1997reasoning}, how to bridge analogy among problems is one of the key abilities in mathematical problem-solving. 
\textcolor{black}{To build an automated MWP solver, we first target to also bridge analogy among problems. Thus we design a module named \emph{analogy identification} for addressing two issues, 1) how to find analogical problems; and 2) how to  bridge the analogy among problems. }

\paragraph{Finding analogical problems} 
Analogical problems usually have associated mathematization processes. In other words, they can be solved by running a similar series of mathematical operators on their own numbers, e.g., the problem 1 and 2 in Figure \ref{fig:intro}.
For analogical problems with different difficulty levels, e.g., the problem 2 and 3 in Figure \ref{fig:intro}, the more difficult one can often be simplified into an easier one. For example, `\emph{buys 4 pencils at $\$0.5$ each}' in problem 3 can be simplified as `\emph{spends $4*0.5=\$2$ on pencils}', which matches well with `\emph{spent $\$4$ to buy pencils}' in problem 2. By this simplification, we can see that the problem 2 and 3 both focus on addressing `number of units = money / unit price'.
To find these analogical problems, we first try to understand how a solution tree is  generated by a decoder. 

Generally, an MWP is solved by a top-down solution tree, where the  nodes closer to the top reflect more about the ultimate goal, i.e., the root of the tree. And lower nodes achieve sub-goals by local calculation. 
Therefore, analogical problems tend to have the same top nodes,
which typically manifest similar required knowledge or solving skills. 
For example, the solution tree of problem 1 and problem 2 in Figure \ref{fig:main} have the same top nodes of running ``/'' and ``-''. 
To find analogical problems, 
we prune a solution tree in a bottom-up manner, removing the bottom nodes and only keeping the top nodes.
Note that we only consider operator nodes because they imply hidden logic \cite{yang2022logicsolver} that different MWPs may share in common, while number nodes represent quantities varying across different MWPs. 
In this way, all solution trees are simplified to have fewer nodes. 
\textcolor{black}{In our experimental implementation, we collect analogical problem pairs if they have the same top-1 operator (the same root node), or the same top-1  and top-2 operators (the same root node and its left children following the pre-order transversal).} 
For an instance, problem 2 and problem 3 in Figure \ref{fig:main} are an analogical  pair (a positive pair) 
since they have common root node ``/'' and its left child ``-''. We consider our operator matching method as a specific kind of measurement of similarity/analogy, incorporating analogies in (a) different topics and (b) different difficulty levels by matching outermost operators in the solution tree. There are two kinds of existing analogy identification methods - 1. Finding problems with similar semantics. 2. Finding problems with similar solutions (trees). However, the former method neglects the topic-topic analogy, and the latter method neglects the difficulty-level analogy.

\paragraph{Bridging the analogy}
\textcolor{black}{For one analogical pair of problems ($X_1$ and $X_2$), they should be mapped close in a representation space, because they share similar knowledge and skills for being solved. To bridge their analogy,  we maximize their analogy score which is defined by}:
\begin{equation}
    s_a = \sigma(MLP([en_x(X_1):en_x(X_2)])),
\end{equation}
where $en_x(X_1)$ yields the  problem representation for $X_1$ by the problem encoder,  
$[:]$ denotes vector concatenation, and $\sigma$ is a sigmod function. \textcolor{black}{The multi-layer perceptrons (MLP) and the encoder $en_x()$ are trained jointly to associate  the  analogical pair. For two non-analogical problems, their analogy score can be calculated in the same way, and should be minimized.}

Then the loss function for bridging the analogy can be defined by cross-entropy:
\begin{equation}
    L_a = -(log ({s_a})+log (1-\overline{s}_a))
    \label{loss_ana_train}
\end{equation}
where $s_a$ is the score for positive pairs and $\overline{s}_a$ is for negative. In each iteration, we randomly sample our positive and negative pairs that are used to update the MLP and encoder. 

\subsection{Solution Discrimination}
Besides the analogical learning part that groups analogical problems together, we also design a solution discrimination module to make the solver associate MWPs stronger to their ground truth solutions. 
\textcolor{black}{We define a discrimination score that measures the association of a problem $X$ with its ground truth solution $Y$:}
\begin{equation}
\label{eq:s_d}
    s_d =  Dis([en_x(X),en_y(Y)]) 
\end{equation}
where $en_y(Y)$ is the representation of equation $Y$ encoded by a gate-recurrent-unit (GRU) \cite{chung2014empirical}.
The discriminator can be empirically performed by a bi-linear similarity function without bias. In other words, the output of $Dis(A,B)$ is $A^TWB$ where $W$ is a learnable matrix.
\textcolor{black}{
For the problem $X$, we can also have a wrong solution (negative sample) $Y^{neg}$. A discrimination score $\bar{s}_d$ can be calculated by Eq.(\ref{eq:s_d}) as well.  Then, this solution discriminator can be trained to minimize ${s}_d$ while maximizing $\bar{s}_d$. In other words, the true solution is pulled close to the problem, while the wrong solution is pushed away from the problem. The discriminator is updated once in each mini-batch. Although the discriminator is trained from scratch, it is stable in converging, as it only encodes solutions where the vocabulary only contains placeholders for numbers and operators.}


\paragraph{Negative Solution Generator}
For   generating negative solutions, 
any manipulation of the ground truth solution can lead to a negative one, as implemented in 
\cite{ijcai2021}. However, the  random modification neglects   the importance of tokens   in the solution equation.  
Although all negative solutions ultimately lead to a wrong answer, the roles they play in minimizing loss functions and serving as contrastive examples to a positive true solution are at different levels of importance. 

Our goal is to find  variants of the ground truth solution as hard negative samples, which only manipulate the most vulnerable (important) token. 
In fact, this target is  similar to evasion attack \cite{carlini2017towards} on texts, i.e., maximum effect and minimum manipulation. Therefore, borrowing the idea from white-box evasion attack, we regard the token with the largest gradient as the most important and vulnerable one:
\begin{equation}
\label{grad}
    y_i = \mathop{argmax\;}_{y_i \in Y}(\nabla Dis([en_x(X),en_y(Y)]).
\end{equation}
After getting the most important $y_i$, the next step is to find the replacement token of it. Different from  \cite{ijcai2021} which  generates it randomly, we notice that the vocabulary for $Y$ is in a small size. Also, the token type has to be consistent after replacement (operator or number). Therefore, it is not costly to find all possible alternatives of token $y_i$ and make them all as negative equations. If $y_i$ is an operator, we have 4 alternatives out of $\{+,-,*,/,\string^\}$, leading to 4 negative equations.  If $y_i$ is a number, the choices of replacement will depend on the size of $V_{num}$ which is usually less than 5. In this way, we make full use of the result of gradient checking and have considered all possibilities of token replacement to form negative solutions.

\paragraph{Additional Negatives} 
The above-generated negative equations are similar to the 
 ground truth solution, since most part of it has not been changed. To generate diverse negative samples that have loose connections to the ground truth, 
we also randomly select a solution from the training set as a negative equation, subject to the vocabulary of that solution being a subset of the current solution vocabulary. Because the output vocabulary may vary in different problems, this restriction ensures the generation of reasonable negative samples. Furthermore, we execute the same procedure on this random negative solution, i.e., finding and manipulating the most vulnerable token to obtain more negative samples. To sum up, the negative equation generator of the solution discriminator module will generate two groups of negative samples which are variants of the ground truth solution and a random solution with its variants. After the discriminator $Dis$ gets well-trained by cross-entropy loss, we can construct a new loss term that provides extra supervision to the encoder of the MWP solver, which is similar to the adversarial loss in the generative adversarial network (GAN) \cite{goodfellow2014generative}:
\begin{equation}
\label{lc}
    L_s = -\log P(s_d = 1|Dis([en_x(X),en_y(Y)])).
\end{equation}
 The solution discriminator thus enhances the association between problem representations and their ground truth equation representations.

\subsection{Model Training}
There are three components in our proposed framework. (a) An analogy identification module that is jointly trained with the encoder-decoder solver, which plays the role of an auxiliary task. (b) A solution discrimination module that is separately trained with the solver, which ensures the consistency between MWP and its true solution, and provides guidance when the encoder-decoder part is trained. (c) The solver is trained by the summation of Seq2Seq loss, analogy identification loss (Equation \ref{loss_ana_train}), and solution discrimination loss (Equation \ref{lc}), where the Seq2Seq loss is the negative log-probability of generating $Y$ given $X$:
\begin{equation}
\label{Seq2Seq_loss}
    L_{seq} = -\log P(Y|X).
\end{equation}
The steps of the training pipeline is in Alg. \ref{alg:algorithm}. Firstly, we randomly sample some positive pairs and negative pairs according to the introduction of analogical identification in Section 3.3. 
Secondly, we calculate the gradient as shown in Equation (\ref{grad}), find all vulnerable tokens, and generate all negative equations for solution discrimination. Then we train the discriminator with a cross-entropy loss with ground truth as the positive sample and manipulated solutions as the negative samples  (line 3-5). Finally, we train the analogy identification module and the solver with the joint loss  (line 6).

\begin{algorithm}

\caption{Training Pipeline}
\label{alg:algorithm}
\textbf{Input}: Problem $X$, Solution $Y$, Analogy Identification Module $\theta$,
Solution Discrimination Module $\gamma$, Encoder-Decoder Module $\delta$\\
\textbf{Parameter}: Loss Weights $\lambda_1, \lambda_2$ \\
\textbf{Output}: Well-trained $\theta$, $\gamma$ and $\delta$\\ \vspace{-0.5cm}
\begin{algorithmic}[1] 
\FOR{$X$, $Y$ in the training set}
\STATE $X_{pos}, X_{neg} \gets$ Sample positive/negative problems from the training set by analogy identification.
\STATE $y_i \gets$ Get the most important token in $Y$ by gradient as shown in Eq. (\ref{grad}).
\STATE $Y_{neg} \gets$ Replace $y_i$ with all possible tokens and get the negative solution set.
\STATE Train the discriminator $\gamma$ with cross-entropy loss.
\STATE Train the analogy identification module (i.e., MLP) $\gamma$ and encoder-decoder module $\delta$ with a joint loss: $L = L_{seq} + \lambda_1 L_a + \lambda_2 L_s$.
\ENDFOR
\STATE \textbf{return} $\theta$, $\gamma$, $\delta$
\end{algorithmic}
\end{algorithm}

\section{Experimental Results}
In this section, we firstly introduce the used datasets, evaluation metrics and baselines in this paper. Then we conduct an accuracy comparison between our solvers and all baseline methods. Next, our ablation studies show the contribution from different modules. We also analyze the solver accuracy under different solution lengths to evaluate generalization ability. Visualization of different groups of MWPs shows that our solver has better MWP representation ability. Case studies can be found in appendix.
\subsection{Datasets}
\paragraph{Math23k}
Math23k \cite{wang2017deep} is the most commonly used Chinese dataset in MWP solving. It contains 23,162 problems with 21,162 training problems, 1,000 validation problems and 1,000 testing problems. We use the value accuracy as the evaluation metric, which checks whether the equation solution given by the model leads to the ground truth value. 
\paragraph{MathQA}
MathQA \cite{amini2019mathqa} is an English mathematical problems dataset at GRE level. The original MathQA dataset is annotated in a different way from Math23k with many pre-defined operations. Also, some problems in the dataset suffer from the low-quality issue. Many efforts \cite{tan2021investigating,li2021seeking,jie2022learning} have been paid to clean and filter the MathQA dataset. In our experiment, we follow the latest version \cite{jie2022learning} of MathQA dataset where 4 arithmetic operators $\{+,-,*,/\}$ are included in this subset. After data filtering, the training set contains 16191 training MWPs and 1605 testing samples. We also use the value accuracy as our evaluation metric for this dataset.

\subsection{Baselines}
The baselines  used in this paper can be divided into RNN-based and PLM-based. For RNN-based solvers, we select GTS \cite{xie2019goal} and Graph2Tree \cite{zhang2020graph} which are the most commonly used baselines in other papers. Besides, we include GTS-teacher \cite{ijcai2021} which pairs up the problems and solutions and checks them with a discriminator, and EEH-G2T \cite{wu2021edge} is the best existing RNN-based solver on Math23k. For PLM-based solvers, we choose the retrieval-based analogical solver REAL \cite{huang2021recall}, contrastive learning solver BERT-CL  \cite{li2021seeking} that determine analogies based on sub-trees in the solution, Deductive Reasoner  \cite{jie2022learning} and Generate\&Rank  \cite{shen2021generate}. To our knowledge, Generate\&Rank is known as the best solver on Math23k (85.6\% accuracy) and Deductive Reasoner performs the best on MathQA (78.6\%).

\begin{table}
\renewcommand\arraystretch{1.12}
\centering
\setlength{\tabcolsep}{1.0mm}{
\begin{tabular}{c|c|c|c}
\hline
            & {Math23k}  & {MathQA} & \#E    \\
\hline
 \multicolumn{4}{c}{RNN-solvers}\\
\hline
GTS         & $75.6$ & $68.0^*$ & $7.2M$        \\
GTS-teacher     & $76.5$ & $68.5^*$ & $7.2M$      \\
Graph2Tree  & $77.4$ & $69.5$ & $9.0M$       \\
EEH-G2T     & $\mathbf{78.5}$ & $-$ & $9.9M$      \\
GTS+A\&D     & $78.2$ & $\mathbf{69.6}$ & $7.2M$      \\
\hline
 \multicolumn{4}{c}{PLM-solvers}\\
\hline
REAL     & $82.3$ & $-$ & $110M$      \\
BERT-CL  & $83.2$ & $73.8$ & $102M$  \\
MWP-BERT  & ${84.6}$ & ${77.2}^*$  & $102M$\\
Deductive Reasoner  & ${85.1}$ & ${78.6}$  & $102M$\\
Generate\&Rank  & ${85.4}$ & $-$ & $610M$  \\
MWP-BERT+A\&D  & $\mathbf{85.6}$ & $\mathbf{79.6}$ & $103M$  \\
\hline
\end{tabular}}
\caption{Comparison of answer accuracy (\%) and the number of parameters in the encoder (\#E). `+A\&D' means that the solver is trained with our proposed analogical pipeline  containing analogy identification and solution discrimination. Accuracy with a `*' indicates  re-produced results by us based on the provided public source code. The other results  come from their papers.}
\label{tab:main_result}
\end{table}

\subsection{Implementation Details}
Our model is trained and evaluated under Pytorch Framework with a single NVIDIA Tesla V100 GPU. The training iteration follows the steps in Alg. \ref{alg:algorithm} where $\lambda_1$ = 0.01 and $\lambda_2$ = 0.001, which are  selected by grid search from [0.0001, 0.001, 0.01, 0.1, 1]. We train the solvers with 160 epochs using a batch size of 32. The learning rate for  RNN-based models and PLM-based models is set to 0.001 and 0.00005, which is halved every 30 epochs. One-epoch training takes about 10 minutes and 30 minutes on GTS-based and BERT-based solvers, respectively. AdamW \cite{kingma2014adam} optimizer is used with default hyper-parameters in Pytorch. Dropout \cite{srivastava2014dropout} probability of 0.5 and 0.1 is used for GTS-encoder and BERT-encoder to avoid potential overfitting. Five-beam search is applied to produce better solutions. For more details, please refer to our code\footnote{\url{https://github.com/LZhenwen/Analogical_MWP}}.

\begin{table*}
\renewcommand\arraystretch{1} 
\centering
\setlength{\tabcolsep}{1.1mm}{
\begin{tabular}{|p{3.8cm}<{\centering}|p{1.5cm}<{\centering}|p{1.5cm}<{\centering}|p{1.5cm}<{\centering}|p{2.3cm}<{\centering}|p{2.3cm}<{\centering}|p{1.3cm}<{\centering}|}
\hline
\multirow{2}{*}{} & \multicolumn{3}{c|}{Analogy Identification (A)} & \multicolumn{2}{c|}{Solution Discrimination (D)} & \multirow{2}{*}{Acc.}\\
\cline{2-6}
& Top-1 MLP & Top-2 MLP & Top-3 MLP & Gradient guided & Extra Negatives &  \\
\hline
MWP-BERT & \xmark & \xmark & \xmark & \xmark & \xmark & $84.7$\\
 & \checkmark & \xmark & \xmark & \xmark & \xmark & $84.9$\\
MWP-BERT+A
           & \checkmark & \checkmark & \xmark & \xmark & \xmark & $ 85.3 $\\
           & \checkmark & \checkmark & \checkmark & \xmark & \xmark & $ 85.1 $\\
           & \xmark & \xmark & \xmark & \checkmark & \xmark & $ 85.0 $\\ 
           & \xmark & \xmark & \xmark & \xmark & \checkmark & $ 85.1 $\\   
MWP-BERT+D         & \xmark & \xmark & \xmark & \checkmark & \checkmark & $ 85.3 $\\   
{MWP-BERT+A\&D} & \checkmark & \checkmark & \xmark & \checkmark & \checkmark & $\mathbf{85.6}$\\
\hline
\end{tabular}}
\caption{Accuracy of different ablated models on the Math23k dataset. `A' denotes analogy identification and `D' denotes solution discrimination. {\checkmark and \xmark \, indicate without/with that module in the ablated models, respectively.}}
\label{tab:ablation1}
\end{table*}

\subsection{Comparative Experiments}
In this part, we compare our model with the above baselines in terms of the value accuracy and the number of parameters in the encoder, as shown in Table \ref{tab:main_result}. On Math23k benchmark, the performance of GTS gets considerably raised from 75.4\% to 78.2\%, which is comparable to the best RNN solver EEH-G2T with fewer parameters in the encoder. For PLM-based solvers, our proposed training method boosts the accuracy of MWP-BERT from 84.6\% to 85.6\%, achieving state-of-the-art performance with much fewer parameters than Generate\&Rank. For MathQA dataset, with  our pipeline, the accuracies of GTS and MWP-BERT both get improved and the improved MWP-BERT solver outperforms all other competitors. For some baselines, we failed to re-produce their methods and get a reasonable accuracy on MathQA, thus we leave them empty as many previous papers \cite{zhang2020graph,li2021seeking,liang2022mwp} did.

\begin{table}
\renewcommand\arraystretch{1.03}
\centering
\setlength{\tabcolsep}{1.5mm}{
\begin{tabular}{|c|c|c|c|c|c|}
\hline
 \multirow{2}{*}{\#OP}&\multirow{2}{*}{Pct.} &\multicolumn{2}{c|}{GTS} &\multicolumn{2}{c|}{MWP-BERT}\\
 \cline{3-6}
 && w/o & w & w/o & w\\
\hline
1 &35.4\%&$84.9$ & $86.1$ &$90.1$ & $\mathbf{90.7}$ \\
2 &44.0\%&$80.6$ & $81.7$ &$87.0$ & $\mathbf{88.0}$\\
3 &14.1\%&$70.7$ & $71.3$ &$62.4$ & $\mathbf{71.3}$\\
4 &3.3\%&$50.0$ & $60.6$&$60.6$ & $\mathbf{78.8}$\\
5\textsuperscript{*} &0.6\%&$38.2$ & $38.2$ &$50.0$ & $\mathbf{50.0}$\\
\hline
\end{tabular}}
\caption{The answer  of BERT and MWP-BERT on problems with different lengths in Math23k.  \#op  shows the No. of operators in the solution. Pct. is the percentage of problems with different number of operators. `w/o' means ``without our 
analogy identification and solution discrimination'', and `w' denotes ``with our method''. * indicates that the result is not statistically meaningful for comparison because there are only 6 samples.} \vspace{-0.3cm}
\label{tab:data_length}
\end{table}
\subsection{Ablation Studies}
There are two major modules in our framework, i.e., analogy identification and solution discrimination. For the former part, we separately test the influence of top-$k (k \leq 3)$ nodes 
and find that only top-1 and top-2 have contributions. The reason for the ineffectiveness of top-3 nodes 
could be that only a small number of MWPs contain more than 3 operators (shown in Table \ref{tab:data_length}), and fewer MWPs would have 3 same operators at the top. Therefore, there will be a considerately limited number of analogical pairs for top-3 MLP. For the solution discrimination module, we evaluate the influence of gradient-guided token selection (we use random token selection in its ablation) and the additional negative solutions that are randomly drawn out of the training set. The results in Table \ref{tab:ablation1} verify the contribution of every individual component to the proposed training pipeline.

\subsection{Analysis of Long-tail Problem Solving}
In Table  \ref{tab:data_length}, the  column Pct. gives the percentage of MWPs with different solution lengths.
We can see that the distribution of MWPs follows a long-tail distribution where the simple MWPs with short solutions are the head, and the difficult MWPs with long solutions are the tail. We   find that the  accuracy on the tails gets more increased than the accuracy on the head with the assistance of our method. Although the overall accuracy improvement on Math23k is about 1\% as shown in Table \ref{fig:main}, our  method makes the solver learn from simple MWPs and generalize to difficult MWPs, bringing a significant accuracy boost on tail 
problems.

\begin{figure}[t!]
\centering 
\includegraphics[width=0.48\textwidth]{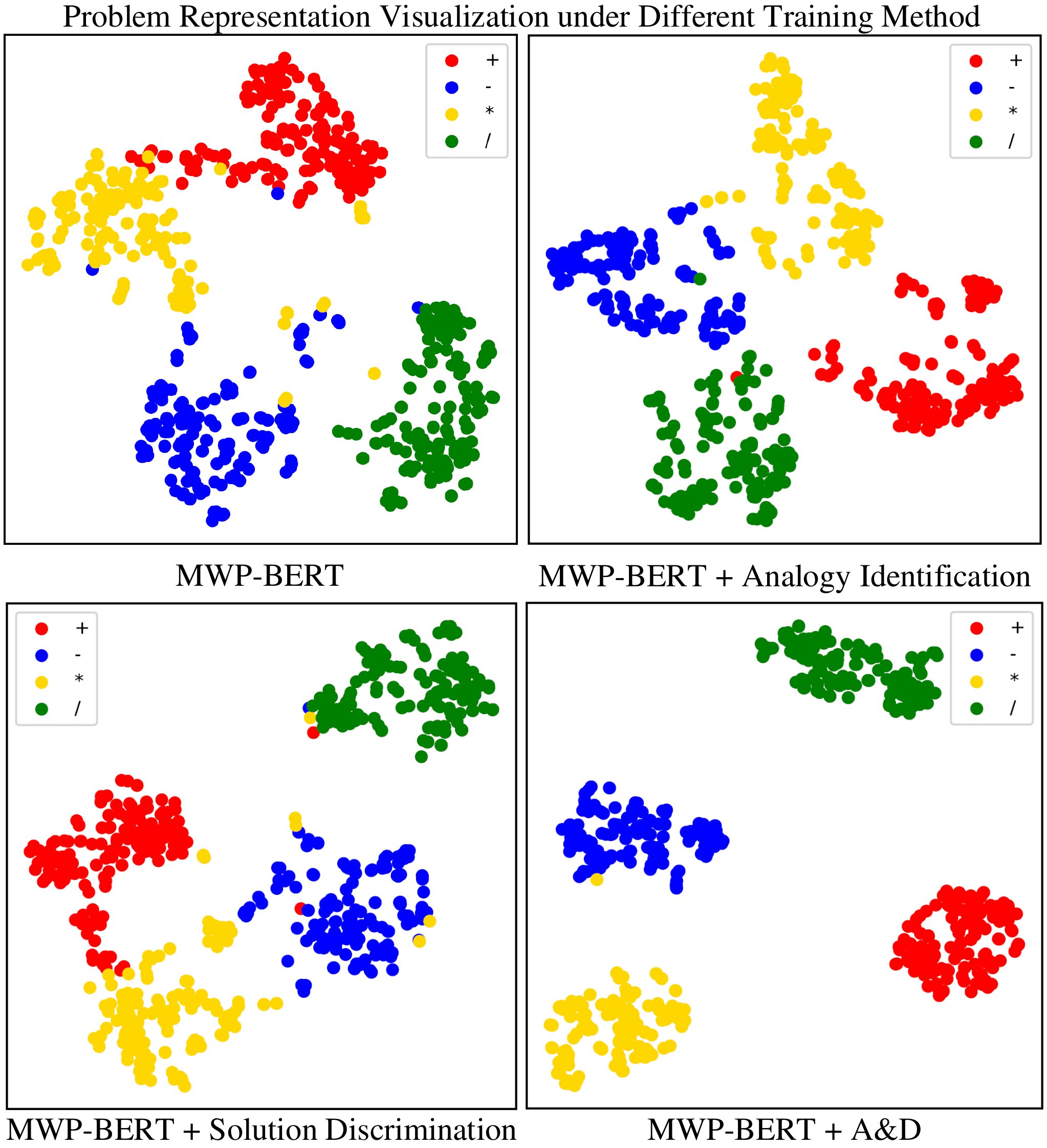}
\caption{Problem representation visualization by T-SNE. Our model with A\&D improves the problem representation learning, which groups analogical problems close and separates non-analogical problems.}
\label{fig:tsne} 
\end{figure}

\begin{figure*}
\centering 
\includegraphics[width=0.925\textwidth]{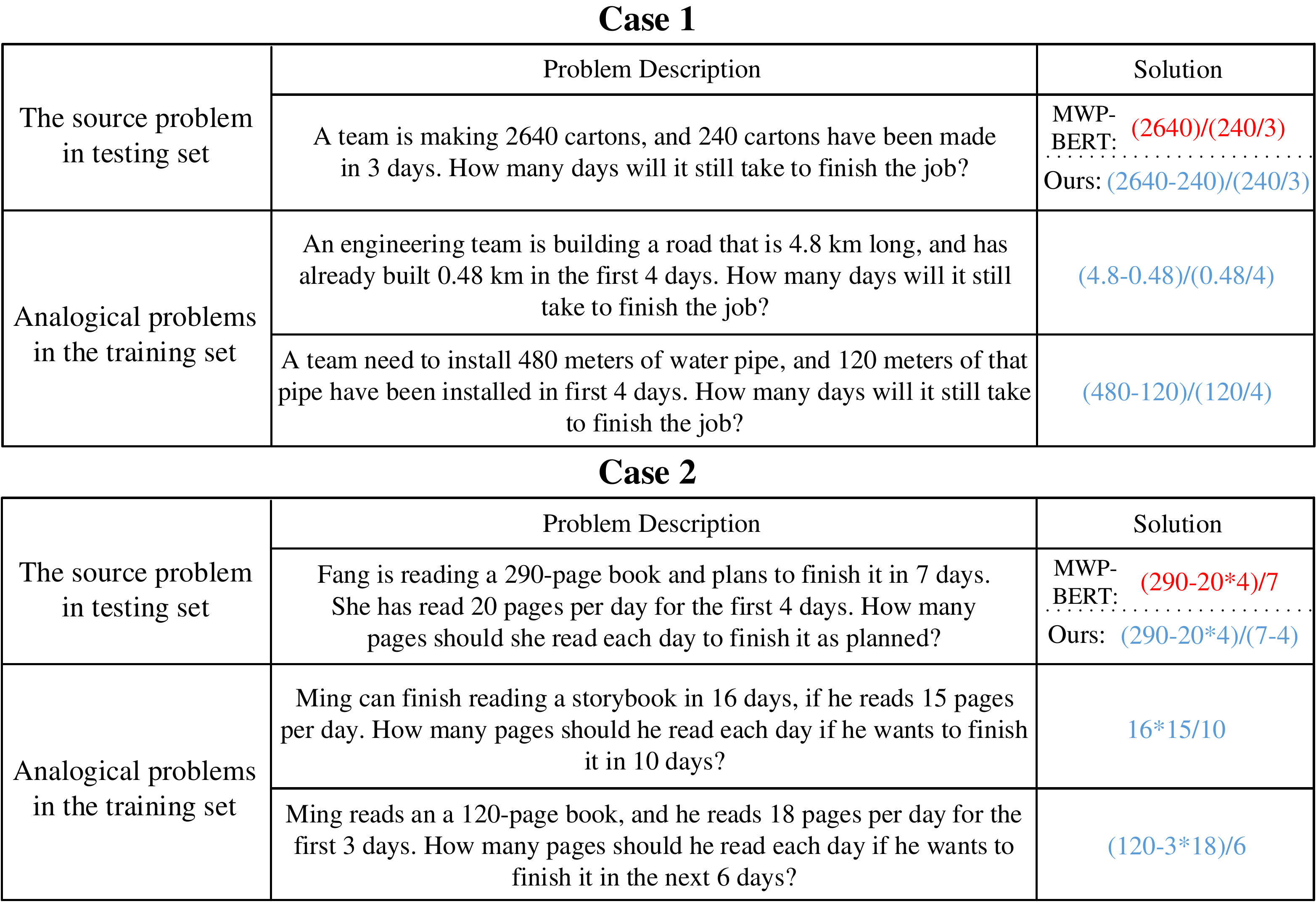}
\caption{Case study.}
\label{fig:case} 
\end{figure*}

\subsection{Visualization}
To demonstrate the ability of grouping analogical problems in  the representation space, we visualize them by T-SNE \cite{van2008visualizing} and check the clustering phenomenon as did in  \cite{ijcai2021,li2021seeking}. We randomly select 150   problems from Math23k in 4 classes, where each class represents a group of MWPs whose solutions has the same root nodes out of $\{+, -, *, /\}$. Then we visualize them  in Figure \ref{fig:tsne}. For vanilla MWP-BERT,  many boundary points are not well separated. With the analogy identification, the number of confusing boundary points gets lower. With the solution discrimination module, the inter-class distance is increased and the green cluster is well separated from the others. With both, different clusters are more separated and the distribution of each cluster gets denser. This analysis  demonstrates that our model effectively improves problem representation learning. 

\paragraph{Case Study}

We show two cases in Figure \ref{fig:case} to demonstrate the potency of our proposed method. We select two problems that MWP-BERT \cite{liang2022mwp} failed to solve from the Math23k testing set. We can find that MWP-BERT solver is influenced by analogical problems in the training set and generates an identical solution as the solution of the second analogical problem. The potential reason is that these MWPs are semantically similar and the model fails to distinguish them. In our approach, we design the solution discrimination module to strength the association between the problems and ground truth solutions. Our solver correctly generates the equation through the help of the designed module.

\section{Conclusion}
We propose a novel analogical training pipeline for math word problem solvers, considering the generalization among analogical MWPs and the association between ground truth solutions and problem descriptions. In this way, problems that need similar solving skills are grouped together to support analogical learning, and solvers are trained to focus on ground truth solutions. The comparative study shows that our method with MWP-BERT outperforms other baselines  on Math23k and MathQA, and is much lighter (with fewer parameters). 
It can also solve more difficult problems due to the analogical reasoning. We believe our work would facilitate the MWP research community and inspire more studies about the analogy in mathematical question answering.

\section*{Limitations}
\paragraph{Commonsense Knowledge} 
As mentioned in \cite{lin2020numersense,DBLP:journals/corr/abs-2107-13435}, MWP solving in the real-word scenario requires many commonsense knowledge, e.g., 1km = 1000m and one day = 24 hours. When these commonsense constants are not explicitly given in the problem description, our MWP solver has no chance to solve problems that require them. A future direction could be injecting commonsense knowledge into MWP solvers.

\section*{Acknowledgement}
The research work is partially supported by the Internal Asia Research Collaboration Grant, University of Notre Dame.

\bibliography{emnlp2022}
\bibliographystyle{acl_natbib}

\end{document}